# Guiding the Sequential Experiments in Autonomous Experimentation Platforms through EI-based Bayesian Optimization and Bayesian Model Averaging

**Ahmed Shoyeb Raihan and Imtiaz Ahmed**

*Department of Industrial and Management Systems Engineering, West Virginia University*
*Morgantown, West Virginia, United States*

## Abstract

Autonomous Experimentation Platforms (AEPs) are advanced manufacturing platforms that, under intelligent control, can sequentially search the material design space (MDS) and identify parameters with the desired properties. At the heart of the intelligent control of these AEPs is the policy guiding the sequential experiments, which is to choose the location to carry out the next experiment. In such cases, a balance between exploitation and exploration must be achieved. A Bayesian Optimization (BO) framework with Expected Improvement based (EI-based) acquisition function can effectively search the MDS and guide where to conduct the next experiments so that the underlying relationship can be identified with a smaller number of experiments. The traditional BO framework tries to optimize a black box objective function in a sequential manner by relying on a single model. However, this single-model approach does not account for model uncertainty. Bayesian Model Averaging (BMA) addresses this issue by working with multiple models and thus considering the uncertainty in the models. In this work, we first apply the conventional BO algorithm with the most popular EI-based experiment policy in a real-life fatigue dataset for steel to predict the fatigue strength of steel. Afterward, we apply BMA to the same dataset by working with a set of predictive models and compare the performance of BMA with the traditional BO algorithm, which relies on a single model for approximation. We compare the results in terms of RMSE and find that BMA performs better than EI-based BO in the prediction task by considering the model uncertainty in its framework.

## 1. Introduction

Recently, with the ever-increasing concerns for sustainability, carbon footprint reduction, less resource and energy consumption, waste minimization, and critical materials preservation, scientists and researchers are more concentrated on a strategic approach for material discovery and development of the manufacturing process. The material design and discovery process tackle the challenge of finding the appropriate compounds, their chemical composition, and processing conditions that satisfy one or multiple target properties [1]. The conventional material discovery process involves iterative trial and error, in which we first select a material design we believe will function well based on prior knowledge; next, we synthesize and test the material in physical experiments; and finally, we use the information we learn from these tests to select the material design to try next. This is a closed-loop process that ends when we find out the optimum target value or exhaust our resources. Such processes, although successful in many scenarios, are not feasible. Discovering materials in this way is expensive and resource-consuming because it requires exploring the Material Design Space (MDS) that comprises all the combinations of the material compositions and processing conditions. Recently, data-driven approaches in material design and discovery are getting a lot of attraction from scientists and researchers. Consequently, Autonomous Experimentation Platforms (AEPs) are emerging, which have all the capabilities to revolutionize the traditional material discovery process [2]–[5]. An AEP is an intelligent system that can independently explore, with little assistance, an MDS of process conditions and material components in search of novel properties or techniques for producing advanced materials [6]. AEPs are inextricably linked to sequential experiment designs where the key is deciding where to conduct the next experiment based on what has been seen thus far. This involves the most important task of finding the balance between exploration and exploitation. The dominant paradigm currently utilized by AEPs is the Bayesian Optimization (BO) framework where this exploration-exploitation trade-off is balanced through the implementation of various acquisition functions such as Expected Improvement (EI). Despite being one of the most significant algorithms in sequential experiment designs, the traditional BO algorithm does not consider the model uncertainty [7]. In material discovery, BO attempts to build a function between the input features (material composition, process conditions etc.) and the target property by constructing a single model that explains this true underlying function. Oftentimes, we do not have any prior knowledge about the true underlying function, and therefore, it is not reasonable to work with a single model. In such cases, a number of potential models or predictive models can be utilized instead of a single model. Each of them has a probability to be the true predictive model that accurately fits the underlying relationship. While the traditional BO approach cannot incorporate this model uncertainty, Bayesian model averaging (BMA) can efficiently address this model uncertainty problem by working with multiple models at a time [7], [8].

In this work, we use a fatigue dataset from the National Institute of Material Science (NIMS) [9]. First, we employ the traditional EI-based BO algorithm to predict the steel fatigue strength, which works with a single predictive model. Later, we build three different predictive models from the same dataset and implement the BMA algorithm, which works with multiple models to predict fatigue strength. We compare the prediction performance of both these approaches using the RMSE metric. Results show that BMA performs better in terms of RMSE by addressing the challenge of resource constraints and dealing with the uncertainty of single models. We design the remaining sections of this work in the following way. Section 2 provides a brief review of the literature on the applications of BO and highlights the importance of incorporating model averaging into the current BO framework. Section 3 discusses the research methodology with the basic concepts of BO and BMA. We discuss the dataset and present our findings in section 4. Finally, in section 5, we point out the limitations of this work and provide the direction for future research in the field of sequential learning.

## 2. Literature Review

Bayesian optimization is a probabilistic model-based technique for the global optimization of black-box functions. It is commonly used in scenarios where the objective function is often time-consuming or expensive to evaluate. In BO, a probabilistic model known as the surrogate model, such as a Gaussian process (GP), is fitted to the objective function based on a limited number of function evaluations. The GP models the uncertainty in the objective function and generates predictions at new points in the input space. Afterward, an acquisition function is used to determine the next sample point to evaluate based on a trade-off between exploration (sampling in regions of high uncertainty) and exploitation (sampling at points with a high predicted improvement over the current best point). This process is repeated until a satisfactory solution is found or a stopping criterion is met. In autonomous experimentation, which deals with sequential experiment designs, BO applications have become very popular due to their capability to accurately model complicated relationships even with very little data [6]. BO is routinely used by researchers in the field of manufacturing, drug and material discovery, neuroscience, pharmacy, and robotics [10]–[13]. To address the issue of determining the ideal additive manufacturing (AM) structure for a particular application, Gongora et al. developed a Bayesian experimental autonomous researcher (BEAR), which combines BO and high-throughput automated experimentation [10]. Burger et al. developed a mobile robot platform driven by the BO algorithm capable of searching for improved photocatalysts [5]. By utilizing the concept of BO to optimize the resampling parameters, Lancaster et al. aimed to increase the accuracy of brain-age prediction [13]. In material design, Zhang et al. incorporated a novel latent-variable (LV) technique in the traditional BO framework to model a mixed-variable GP [12]. While BO actively eliminates the need to conduct many experiments, it cannot deal with model uncertainty. To deal with this challenge, Talapatra et al. introduce BMA in the EI-based BO framework [7]. The authors argue that their approach is more effective because instead of assuming beforehand what the best predictive model is and iteratively updating it based on limited initial data, they deal with model uncertainty by considering a range of potential predictive models. Through BMA, although we have to work with a set of predictive models instead of a single model, it is found that as more sequential experiments are performed, only one model, which is the true predictive model, gets a higher probability of getting selected. As this process is taking place, the optimum target properties of the material are also being identified simultaneously.

## 3. Research Methodology
### 3.1. Bayesian Optimization
Bayesian optimization is a sample-efficient and probabilistic approach for the global optimization of black-box functions. It combines a probabilistic model of the objective function with an acquisition function to balance exploration and exploitation in the search for a high-performing solution. As the probabilistic model, also known as the surrogate function, we use GP in the BO framework because of its flexibility and adaptivity. A GP is a statistical method where a random variable $f(x)$ is assigned to each point $x \in \mathbb{R}^P$ (with $P$ being the number of parameters), and the finite collection of these random variables follows a multivariate Gaussian distribution which is:

$$Gp \sim p(f|X) = N(f|\mu, K) \quad (1)$$

In the above equation, $\mathbf{f} = \{f(x_1), \ldots, f(x_n)\}$, $\boldsymbol{\mu} = m(X) = \{m(x_1), \ldots, m(x_n)\}$ is the mean function and $\mathbf{K} = k(x_i, x_j)$ is the covariance function. For the acquisition function, we use EI. With GP as a surrogate model and EI as an acquisition function, BO attempts to identify the optimal inputs, $x_{opt}$, to determine the global maximum (or minimum) of the MDS. The following equation represents this:

$$x_{opt} = arg \max_{x \in \mathbb{R}^P} f(x) \quad (2)$$

The acquisition function EI, which decides where to conduct the next experiment, is expressed as:

$$EI_n(x) = (\mu_*(x) - f(x^+))\Phi\left(\frac{\Delta_n(x)}{\sigma_*(x)}\right) + \sigma_*(x)\phi\left(\frac{\Delta_n(x)}{\sigma_*(x)}\right) \quad (3)$$

In equation (3), $\Delta_n(x) = \mu_*(x) - f(x^+)$ captures the potential improvement over the current best solution $(x^+)$ while $\mu_*(x)$ and $\sigma_*(x)$ denote mean and standard deviation of the GP posterior predictive at $x$, respectively. A flowchart with the EI-based BO algorithm is shown in Figure 1(a).

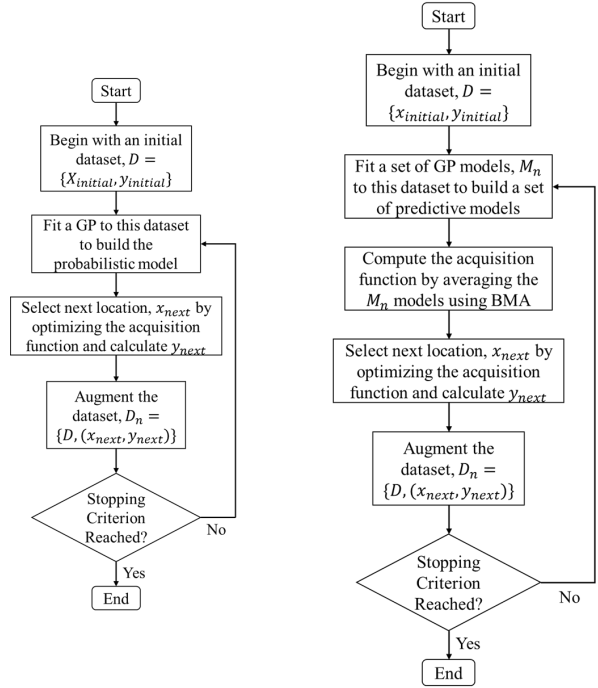

Figure 1: Flowcharts for the algorithms: (a) EI-based BO algorithm (left) (b) BMA algorithm (right)

### 3.2. Bayesian Model Averaging

In sequential experiment design, BMA is a technique for uncertainty quantification, where multiple models are averaged to predict the target output. The algorithm of BMA proposed in this work is somewhat similar to the algorithm of the EI-based BO. The difference is that, in BMA, as we are working with a set of statistical models (for example, a set of GPs), there is a model averaging step where the potential models are averaged, depending on the posterior model probabilities, and the acquisition function is computed and consequently optimized to get $x_{opt}$ with model averaging. The predictive probabilistic model of y for a new feature vector x after observing the data $D$ is:

$$P((y|x, D) = \sum_{i=1}^{L} P(M_i|D)P(y|x, D, M_i) \quad (4)$$

In this equation, $P(y|x, D, M_i)$ represents each potential probabilistic model $M_i$. The posterior model probabilities for each model are calculated using the following equation:

$$P(M_i|D) = \frac{P(D|M_i)P(M_i)}{\sum_{j=1}^{L} P(D|M_j)P(M_j)} \quad (5)$$

Here, $L$ represents the total number of models under consideration. A flowchart with the steps of the BMA algorithm is shown in Figure 1(b).

## 4. Fatigue Life Dataset: A Case Study

### 4.1. Dataset

The fatigue dataset used in this study to predict the fatigue strength of steel was obtained from the National Institute of Material Science (NIMS). The pre-processed dataset has 437 samples, i.e., 437 previously conducted experiments. The initial dataset has about 25 features which are in turn divided into four classes: chemical composition (percentage of carbon, chromium, silicon, nickel, etc. present in the alloy), upstream processing details (ingot size, reduction ratio, etc.), heat treatment methods (normalizing, quenching, hardening, carburizing, etc.) and mechanical properties (yield strength, ultimate tensile strength, fatigue strength, etc.). For the sake of simplicity in this work, we use 14 features from the chemical composition and methods of heat treatment class. The target property is the fatigue strength of steel. A description of the features is shown in Table 1.

Table 1: Input features used to predict the fatigue strength of steel

| Feature | Description | Feature | Description |
|---|---|---|---|
| NT | Normalizing Temperature | TCr | Cooling Rate for Tempering |
| THT | Through Hardening Temperature | C | % of Carbon |
| THQCr | Cooling Rate for Through Hardening | Si | % of Silicon |
| CT | Carburization Temperature | Mn | % of Manganese |
| DT | Diffusion Temperature | Ni | % of Nickel |
| QmT | Quenching Media Temperature | Cr | % of Chromium |
| TT | Tempering Temperature | Mo | % of Molybdenum |

In Table 1, the features NT, THT, THQCr, CT, DT, QmT, TT, and TCr belong to the heat treatment methods category whereas C, Si, Mn, Ni, Cr, and Mo belong to the chemical composition category.

### 4.2. Evaluation Metric (RMSE)

To compare the prediction performance of the fatigue strength of steel, we use the root mean squared error (RMSE) as the evaluation metric. It is expressed using the following equation:

$$RMSE = \sqrt{\frac{1}{N}\sum_{i=1}^{i=N}(y_i - \hat{y}_i)^2} \qquad (6)$$

In the above equation, $N$ is the size of the test dataset, $\hat{y}_i$ is the predicted value of the model, $y_i$ is the true response value. The lower the RMSE value, the better the model's prediction performance.

### 4.3. Results from EI-based BO

At first, we employ the traditional EI-based Bayesian Optimization and calculate the RMSE value to get an overview of the prediction performance. This is a single model-based approach. We use all the input features (see Table 1) to create a relationship between these features and the target property (fatigue strength) using GP as the statistical model and EI as the acquisition function. We start with 5 initial experiments and add 3 experiments to this starting dataset at every iteration based on the optimization of the acquisition function. We keep our budget fixed to 125 experiments indicating that we perform a total of 40 iterations. This suggests that from the original 437 experimental data points, we use 5 for the initial and 120 for the sequential experiments. The model is thus trained on these 125 points. Later, we use the remaining points to create the test dataset, where we apply RMSE to measure the performance of the trained model. To quantify the underlying uncertainty in RMSE, we repeat the above process 20 times. The results are shown in the following table.

Table 2: Performance of the EI-based BO for 20 runs using the RMSE metric

| Run | 1 | 2 | 3 | 4 | 5 | 6 | 7 | 8 | 9 | 10 |
|---|---|---|---|---|---|---|---|---|---|---|
| RMSE | 73.32 | 251.53 | 276.61 | 2263.57 | 808.27 | 193.53 | 32.99 | 171.88 | 226.61 | 165.25 |
| Run | 11 | 12 | 13 | 14 | 15 | 16 | 17 | 18 | 19 | 20 |
| RMSE | 145.94 | 163.21 | 171.16 | 695.56 | 1389.03 | 517.57 | 337.68 | 508.39 | 269.15 | 147.23 |

### 4.4. Results from BMA

The EI-based BMA is the second approach we propose in this study to predict the fatigue strength of steel. As mentioned in section 3.2, we need a set of predictive models in BMA, which are then averaged, and the acquisition function is optimized based on the individual weights of the model. We create 3 different models from all the 14 features with 6 features in each. The first model consists of only the features from the heat treatment method class.

The second model contains features from the chemical composition class. The third and last model comprises 6 features from both categories. The three models with their features are mentioned in the following table.

Table 3: List of the three models with their respective features for BMA

| Model Number | Features in the Model |
|---|---|
| Model 1 | [NT, THT, THQCr, DT, TT, TCr] |
| Model 2 | [C, Si, Mn, Ni, Cr, Mo] |
| Model 3 | [QmT, CT, NT, C, Ni, Cr] |

As with the EI-based BO, in BMA, we start with 5 initial experiments and perform a total of 40 iterations, resulting in 120 sequential experiments. This indicates that we are adding 3 data points sequentially at every iteration. After training the BMA model on these 125 experiments, we test the prediction capability using RMSE. The entire process is repeated 20 times to get an idea of the uncertainty in RMSE. This same approach already used in the EI-based BO allows us to get an accurate and unbiased comparison between BO and BMA. The results from BMA are shown in the following table:

Table 4: Performance of BMA for 20 runs using the RMSE metric

| Run | 1 | 2 | 3 | 4 | 5 | 6 | 7 | 8 | 9 | 10 |
|---|---|---|---|---|---|---|---|---|---|---|
| RMSE | 144.99 | 112.49 | 90.19 | 232.96 | 164.47 | 87.94 | 172.12 | 131.40 | 64.29 | 165.47 |
| Run | 11 | 12 | 13 | 14 | 15 | 16 | 17 | 18 | 19 | 20 |
| RMSE | 98.69 | 109.90 | 125.34 | 249.65 | 101.32 | 71.75 | 123.44 | 97.62 | 226.90 | 84.38 |

In BMA, the 3 models we initially constructed are averaged, and their weights are updated every iteration. Consequently, the model that best describes the true underlying function has a larger weight than the other models. We determined the weights of these 3 individual models in BMA for the 20 runs. The results for the first 10 are presented in the following table.

Table 5: Weights of the 3 models in BMA for 20 runs

| Run | 1 | 2 | 3 | 4 | 5 | 6 | 7 | 8 | 9 | 10 |
|---|---|---|---|---|---|---|---|---|---|---|
| Model1 | 0.550 | 0.521 | 0.525 | 0.506 | 0.475 | 0.550 | 0.547 | 0.475 | 0.561 | 0.478 |
| Model2 | 0.182 | 0.189 | 0.202 | 0.225 | 0.270 | 0.049 | 0.200 | 0.089 | 0.118 | 0.254 |
| Model3 | 0.268 | 0.290 | 0.274 | 0.269 | 0.255 | 0.401 | 0.254 | 0.436 | 0.320 | 0.268 |

The results show that the mean weight of Model 1, Model 2, and Model 3 from the 20 runs is 0.537, 0.166, and 0.297, respectively, suggesting that Model 1 is possibly the best model describing the true underlying function.

### 4.5. Performance Comparison of BO and BMA

To visualize the performance of BO and BMA, we construct a boxplot where RMSE is used as the performance metric. This is illustrated in Figure 1. It is seen that BMA performs better than BO by not only having a lower RMSE value but also having a significantly smaller deviation from the mean of the RMSE values. The mean RMSE values for BO and BMA after 20 runs are found to be 440.424 and 132.767, respectively. The standard deviation of the RMSE values for these 20 runs is 533.825 for BO and 53.985 for BMA.

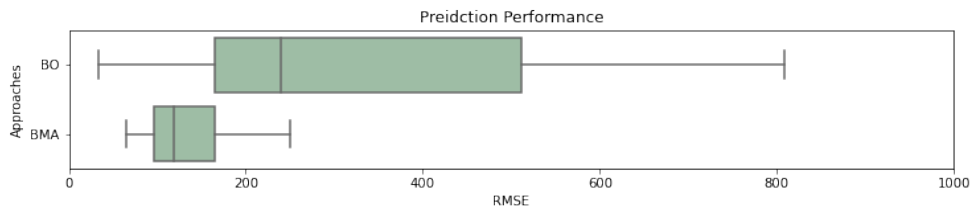

Figure 2: Performance comparison of BO and BMA in terms of RMSE

## 5. Conclusion

In this study, we applied BO and BMA based on the EI-based acquisition function to predict the fatigue strength of steel. We found that the performance of BMA based on multiple models was significantly better than the single model-based BO widely used as the decision maker in developing the AEPs. The single model-based BO, despite having more features in the model, was outperformed by BMA where each model was comprised of a few features. This suggests that including more features in a model can deteriorate the performance of that model. We also found that in BMA, the models' weights get updated as more experiments are carried out, and the true predictive model gets more importance in the long run. Indeed, when we performed an EI-based BO on the best model (Model 1), we found that its performance is very close to BMA in terms of RMSE. Working with multiple models in BMA not only eliminates model uncertainty but also finds the best model for us. This study has some limitations. The EI-based approach used in this work is considered greedy, favoring exploitation over exploration in some literature. In addition, the initial selection of the 3 models in BMA was made only through a brief review of the existing literature. This might have resulted in higher RMSE values in BMA, although lower than the single model-based approach. Future research on BMA could involve using other acquisition functions, including surprise-reactive experiment policies, which are argued to have a better and quick approximation property. An improved feature selection strategy with a thorough literature review can be applied to create the initial models for BMA. The number of initial experiments can be varied along with the number of experiments performed at every iteration to check if there is any change in the performance.

## Acknowledgments

The authors are grateful to the Department of Industrial and Management Systems Engineering of West Virginia University for the support in completing this work.